\documentclass[12pt,letterpaper]{article}
\usepackage{geometry}
\usepackage[utf8]{inputenc}

\usepackage{float}

\usepackage{cite}

\usepackage{nameref,hyperref}

\usepackage{bookmark}
\usepackage{graphicx}%
\usepackage{multirow}%
\usepackage{amsthm}%
\usepackage{mathrsfs}%
\usepackage[title]{appendix}%
\usepackage{xcolor}%
\usepackage{textcomp}%
\usepackage{manyfoot}%
\usepackage{booktabs}%
\usepackage{algorithm}%
\usepackage{algorithmicx}%
\usepackage{algpseudocode}%
\usepackage{listings}%
\usepackage{amssymb}
\usepackage{amsmath}

\DeclareMathOperator*{\argmin}{arg\,min}

\usepackage{changepage}

\usepackage[aboveskip=1pt,labelfont=bf,labelsep=period,singlelinecheck=off]{caption}

\makeatletter
\renewcommand{\@biblabel}[1]{\quad#1.}
\makeatother

\usepackage{lastpage,fancyhdr,graphicx}
\usepackage{epstopdf}
\fancyheadoffset[L]{2.25in}
\fancyfootoffset[L]{2.25in}

\usepackage{color}

\definecolor{Gray}{gray}{.25}

\usepackage{graphicx}

\begin{document}
\vspace*{0.35in}

\begin{flushleft}
{\Large
\textbf\newline{Mapping minds not averages: a scalable subject-specific manifold learning framework for neuroimaging data}
}
\newline
\\
Eloy Geenjaar\textsuperscript{1, 2},
Vince Calhoun\textsuperscript{1, 2}
\\
\bigskip
\bf{1} Electrical and Computer Engineering, Georgia Institute of Technology
\\
\bf{2} (TReNDS) Translational Research in Neuroimaging \& Data Science center, Georgia State University, Georgia Institute of Technology,  \& Emory University
\\
\bigskip
egeenjaar@gatech.edu

\end{flushleft}

\section*{Abstract}
Mental and cognitive representations are believed to reside on low-dimensional, non-linear manifolds embedded within high-dimensional brain activity. Uncovering these manifolds is key to understanding individual differences in brain function, yet most existing machine learning methods either rely on population-level spatial alignment or assume data that is temporally structured, either because data is aligned among subjects or because event timings are known. We introduce a manifold learning framework that can capture subject-specific spatial variations across both structured and temporally unstructured neuroimaging data. On simulated data and two naturalistic fMRI datasets (Sherlock and Forrest Gump), our framework outperforms group-based baselines by recovering more accurate and individualized representations. We further show that the framework scales efficiently to large datasets and generalizes well to new subjects. To test this, we apply the framework to temporally unstructured resting-state fMRI data from individuals with schizophrenia and healthy controls. We further apply our method to a large resting-state fMRI dataset comprising individuals with schizophrenia and controls. In this setting, we demonstrate that the framework scales efficiently to large populations and generalizes robustly to unseen subjects. The learned subject-specific spatial maps our model finds reveal clinically relevant patterns, including increased activation in the basal ganglia, visual, auditory, and somatosensory regions, and decreased activation in the insula, inferior frontal gyrus, and angular gyrus. These findings suggest that our framework can uncover clinically relevant subject-specific brain activity patterns. Our approach thus provides a scalable and individualized framework for modeling brain activity, with applications in computational neuroscience and clinical research.

\clearpage
\section{Introduction}\label{sec:introduction}
Despite the universal nature of human brain function, brain dynamics that underpin cognition are uniquely expressed in each person's brain. It has become increasingly clear that the brain activity that generates these cognitive functions can be expressed on a low-dimensional manifold~\cite{iyer2022focal}. Neuroimaging studies have highlighted that although functional magnetic resonance imaging (fMRI) data can be expressed in a low-dimensional manifold~\cite{gao2021nonlinear,song2023large,busch2023multi}, its representational geometry varies between subjects~\cite{charest2014unique,charest2015brain} due to differences in functional activation maps. Existing methods such as shared response modeling (SRM)~\cite{chen2015reduced,nastase2019measuring,richard2019fast,richard2020modeling} and hyperalignment (HA)~\cite{haxby2020hyperalignment} address variation in individual functional activation maps by allowing subject-specific spatial variability while assuming consistent temporal dynamics or shared information, respectively. In contrast, atlas/region of interest (ROI)-based approaches assume consistent group-level regions across subjects, but allow subject-specific temporal variations. Data-driven approaches such as group independent component analysis (ICA) focus on capturing spatio-temporal variation by fitting a group model and estimating subject-specific maps and timecourses via back-reconstruction~\cite{calhoun2001method}. Such models have been shown to capture subject-specific variability well in both spatial and temporal dimensions~\cite{allen2012capturing} while allowing for correspondences among subjects within the context of a linear mixing model. Motivated by this, our framework focuses on learning non-linear temporal manifolds, while capturing both subject-specific and correspondences among subjects' spatial and temporal information.

Using machine learning to find low-dimensional manifolds from fMRI data has improved the quality of those manifolds. In fact, recent work has shown that subject-specific manifolds can greatly improve the prediction performance of events in a movie~\cite{busch2023multi}. It is, however, harder to compare subjects when each is projected onto their own manifold, instead of a shared group manifold. Moreover, it is challenging to project unseen timepoints onto the learned manifold with the proposed method. On the other hand, neural network approaches, which can generalize to new timepoints and subjects rather well, require a larger amount of training data. Neural networks are thus often trained on datasets with many subjects, and it is implicitly assumed that spatial variations across subjects can be captured by the non-linearities in the neural networks~\cite{huang2017modeling,han2019variational,kim2021representation}. To force the encoder to capture subject-specific features, a recent approach uses subject-specific decoders to learn a latent manifold~\cite{huang2022learning}. Training these separate subject-specific decoders is computationally expensive, making it hard to scale to a large number of subjects. A more efficient method, which only uses a separate linear layer for each subject, was used to train a shared visual decoding model from fMRI data~\cite{scotti2024mindeye2}. It is not clear whether this type of approach works well for low-dimensional manifold learning, and although it is more efficient, it still does not scale well to a large input size and a large number of subjects, see Appendix~\ref{app:scaling}. 

In this work, we therefore look at incorporating subject-specific spatial variation for non-linear manifold learning. We both propose a model that uses a separate linear layer for each subject, and a parameter-efficient model that scales to a large input size and number of subjects. First, we show that linear subject-specific maps improve non-linear manifold learning in Section~\ref{sec:reconstruction-results}. Second, we show that our parameter-efficient model can be trained on a large number of subjects, and generalize to unseen subjects in Section~\ref{sec:subject-generalization}. Moreover, we find that subject-specific weights are different for control subjects than for schizophrenia subjects, and that these differences also generalize to a group of unseen subjects. Lastly, we find interpretable spatial group differences between control subjects and subjects diagnosed with schizophrenia in Section~\ref{sec:schizophrenia-results}.

\begin{figure}[H]
    \centering
    \includegraphics[width=\textwidth]{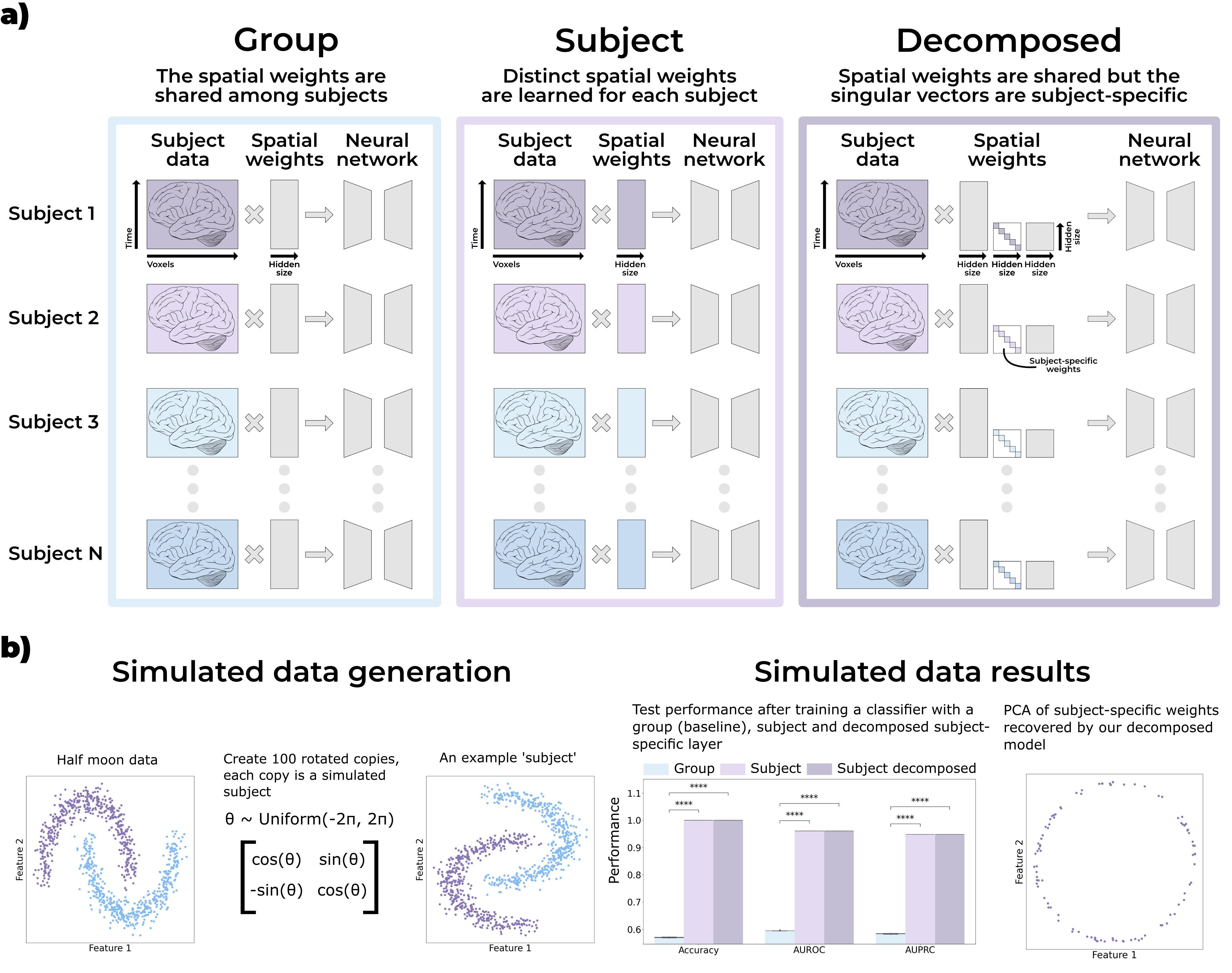}
    \caption{\textbf{Subfigure a}: The three main models tested in our framework, subject-specific parameters have different colors across subjects, and parameters shared across subjects are gray. The Group model assumes the same spatial weights and neural network for all subjects in a dataset, and requires the smallest number of parameters. This is currently the most common way neural networks are applied to neuroimaging data. The Subject model assumes a different spatial weights matrix for each subject, but shares the same neural network across the subjects in the dataset. This model requires the largest number of parameters, especially with many subjects or large spatial inputs. The Decomposed model assumes a matrix decomposition, and assumes that only the singular values are different for each subject. This greatly reduces the number of parameters used compared with the Subject model, and allows it to both capture subject-specific spatial maps and scale to a large number of subjects or large spatial maps. \textbf{Subfigure b}: The simulated data experiment. Each simulated subject is a randomly rotated version of the half moon dataset. We train a multi-layer perceptron (MLP) using the three different models, and show that the Group model can not capture the subject differences, even when a neural network is used. The Subject and Decomposed models achieve almost perfect performance, and the singular values of the Decomposed model projected onto the first two principal components form a circle. This circle corresponds to the different subjects, and clearly captures the rotational generation process (the unit circle).}
    \label{fig:method}
\end{figure}

\begin{figure}[t]
    \centering
    \includegraphics[width=\textwidth]{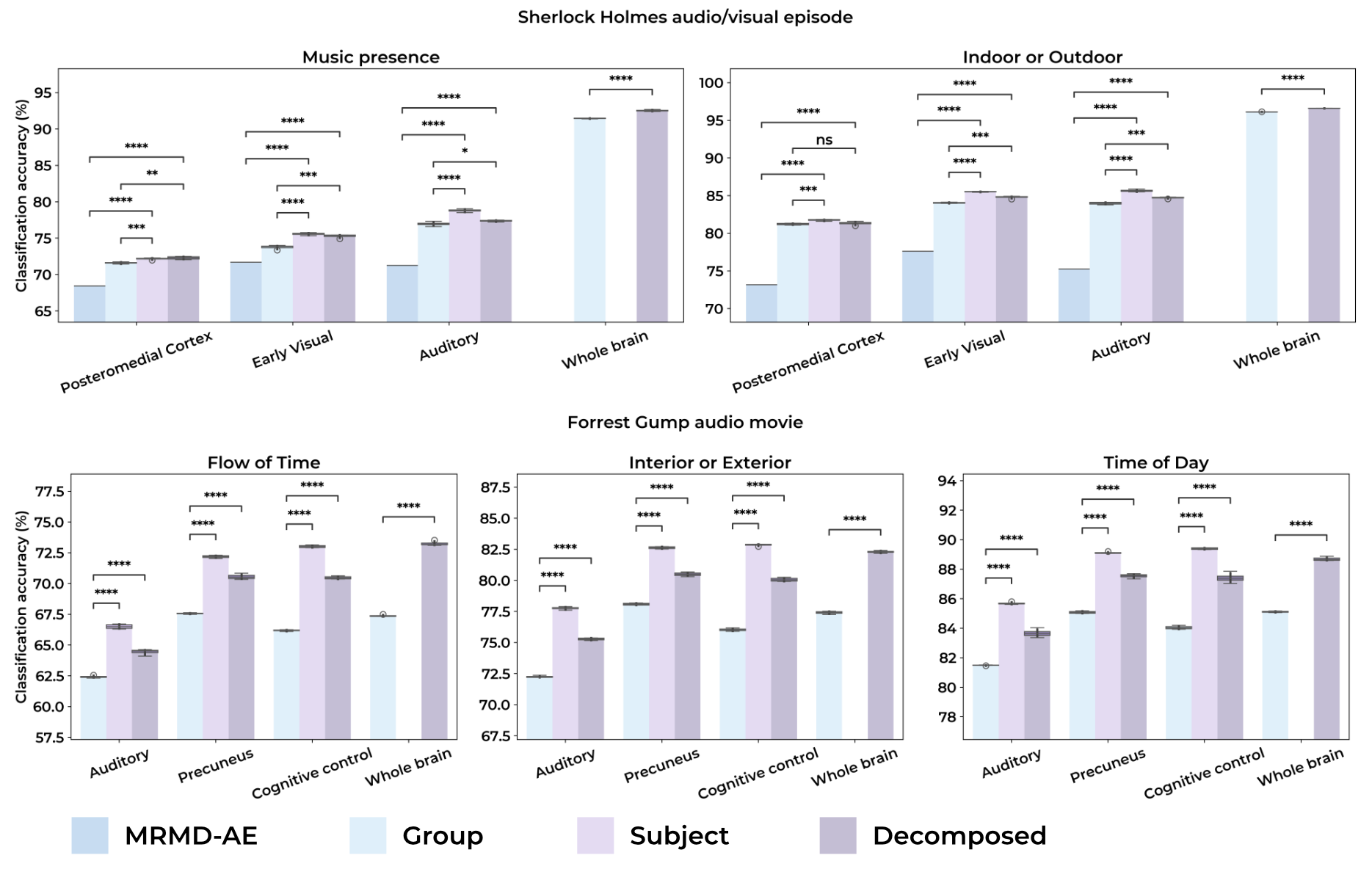}
    \caption{The classification results in this figure are calculated with a radial basis function (RBF) support vector machine (SVM) for auto-encoder models that have embedded an unseen 50\% of the timeseries. The top row shows performance for the auditory, early visual, posteromedial cortex regions of interest (ROIs), and whole-brain data from the Sherlock dataset~\cite{chen2017shared}. Separate RBF-SVMs are trained for the music presence (binary) and indoor vs outdoor (binary) classification. The bottom row shows the performance for the auditory, precuneus, and cognitive control ROIs, using whole-brain data from the auditory Forrest Gump dataset~\cite{hanke2014high}. The three classification types are flow-of-time (4-way), whether a scene is interior or exterior (binary), and time-of-day (binary). In almost all cases, the Subject and Decomposed models significantly outperform the Group and manifold-regularized multiple decoder, autoencoder (MRMD-AE) models. The results for the MRMD-AE are taken from the original paper~\cite{huang2022learning}. Especially for whole-brain data, classification performance is high and significantly better, but the Subject model can not be evaluated in this setting due to bad memory scaling.}
    \label{fig:classification}
\end{figure}

\section{Experiments}\label{sec:experiments}
In our work we compare three main neural network types, which are visualized in Figure~\ref{fig:method}a. The baseline Group model shares the first linear layer in the neural network across all of the subjects in the dataset. The Subject model trains a separate linear layer for each subject in the dataset. Lastly, the Decomposed model shares much of the weights in the first layer, but trains separate singular vectors for each subject in the dataset.

\subsection{Subject-specific simulations}
\label{sec:simulation-results}
The goal of the simulation experiments is to test that the Group model, even with a large number of non-linear layers, can not recover the true labels of a simple toy dataset when we generate subjects with linear rotations. As discussed in the introduction, it is often an implicit assumption when applying neural networks on fMRI data that the non-linearities can capture the subject-specific differences. In the simulation experiment, we show that this may not necessarily be true, and that both our Subject and Decomposed models can easily capture the underlying distribution with almost perfect accuracy.

To ensure a specific hyperparameter setting is not the reason the Group model can not recover the true labels, we evaluate 24 hyperparameter settings across four seeds and pick the best average hyperparameters for each of the three models on a validation set. Similar to naturalistic stimulus fMRI datasets, we assume a label exists for each timestep. Specifically, we simulate the timeseries for each subject as 1000 samples from the half moons dataset~\cite{scikit-learn} with a $\sigma=0.1$ noise level and rotate the 1000 samples with 100 random angles $\theta \sim \mathcal{U}\left[-2\pi, 2\pi \right]$ to generate 100 subjects. We then compare a multi-layer perceptron (MLP) classifier as the neural network in our Subject and Decomposed models to the same MLP in the Group model, see Figure~\ref{fig:method}a. The results in Figure~\ref{fig:method} indicate that the Group model, which is generally how neural networks are trained on fMRI data, does not capture subject-specific differences with its non-linearities. However, an MLP coupled with our Subject and Decomposed models achieves an almost perfect classification accuracy. In Figure~\ref{fig:method}b we additionally show that the Decomposed model exactly recovers the underlying subject generation process. Specifically, each subject corresponds to an angle between $-2\pi$ and $2\pi$, and our layer captures this structure as a circle in 2 dimensions. This exactly represents the underlying generation process because the unit circle is a representation of all possible angles between $-2\pi$ and $2\pi$. The subject-specific weights the Decomposed model learns thus provide an intuitive interpretation of subject differences.

\begin{figure}[t]
    \centering
    \includegraphics[width=\textwidth]{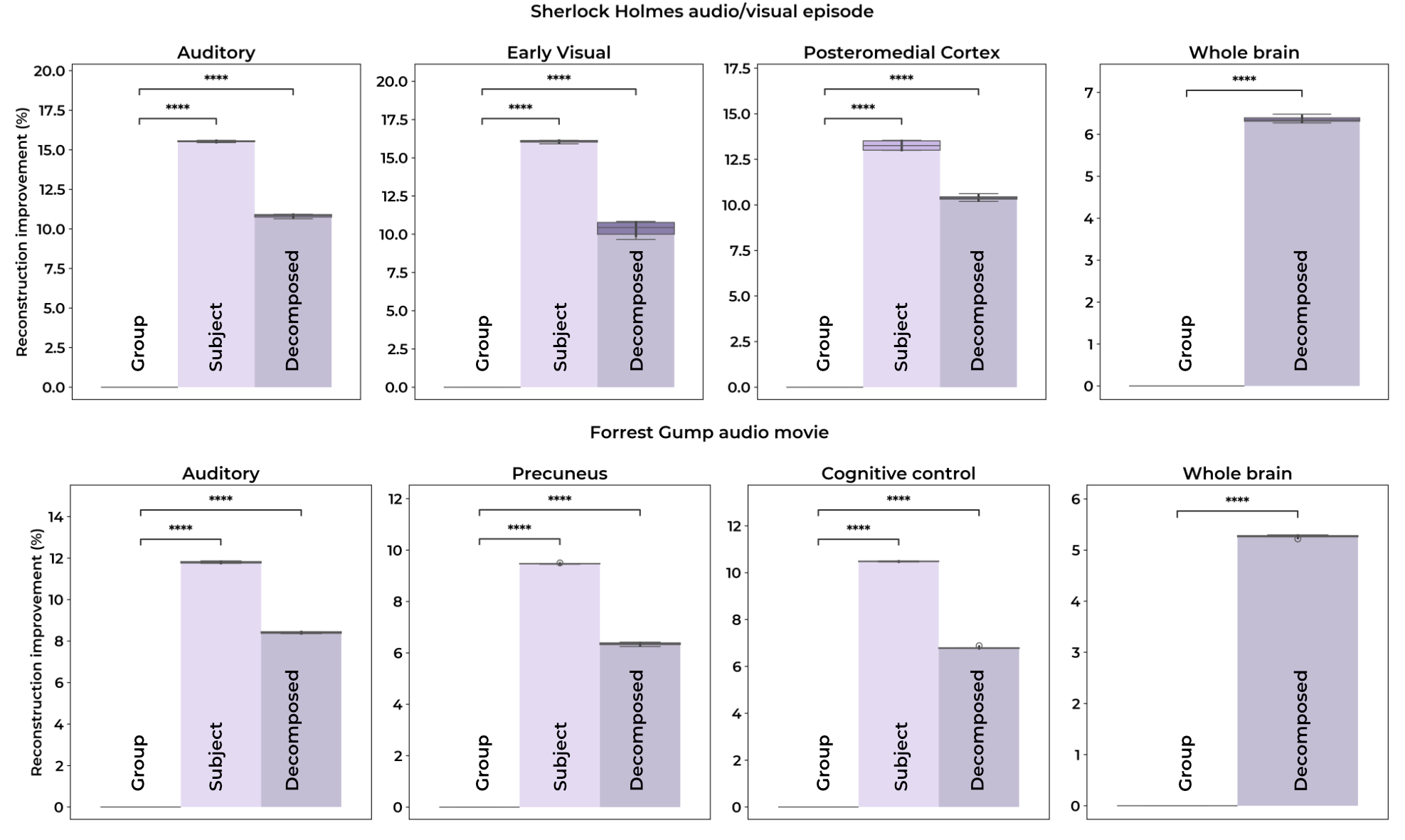}
    \caption{Reconstruction improvements in terms of percentage reduction in mean squared error over the Group model on the test set. For both datasets and each region of interest (ROI) or whole-brain data, both the Subject and Decomposed models significantly outperform the Group model in terms of generalization to new fMRI data from the same subjects. This indicates that using subject-specific spatial maps helps the autoencoder capture more of the variance in the data in a generalizable manner.}
    \label{fig:autoencoding}
\end{figure}

\subsection{Auto-encoding manifold evaluation}
\label{sec:reconstruction-results}
Naturalistic fMRI datasets help us understand how complex stimuli affect brain signals and what dynamical signatures correspond to complex cognitive processes in the brain. Additionally, the context of the naturalistic stimuli at each timestep is often used as a label to help evaluate fMRI methods. For example, a label may indicate whether a scene in a movie is currently taking place outside, whether music is playing in the background, or what time of day it is at a particular point in a story. These labels are then used in a classification task to evaluate the quality of the manifold that is learned for each subject~\cite{busch2023multi}, for example. Given the relationship between naturalistic stimuli and complex cognitive processes, fMRI analysis methods that can better capture these labels will also be able to accurately capture complex cognitive processes. Since detecting deviating cognitive processes is important for mental disorder research, methods that better capture cognitive processes using fMRI are thus useful for studying mental disorders.

Since many datasets, such as resting-state fMRI, do not have timestep-level labels, we evaluate how well our layer performs as part of a neural network that does not assume any structure in the data. We use the same evaluation method as the MRMD-AE~\cite{huang2022learning} and train an autoencoder on the first half of the Sherlock~\cite{chen2017shared} and 7T audio version of the Forrest Gump naturalistic fMRI datasets~\cite{hanke2014high}. The second half of the fMRI timeseries is our test set. After embedding the test set with our trained autoencoder, we use the test set's low-dimensional embeddings to perform a 5-fold classification task on the timestep-specific context labels. This classification task evaluates how well each model generalizes to new fMRI data from the same subject in terms of capturing cognitive processes. Similar to the simulation results, we use our Decomposed, Subject, and Group models in MLP-based autoencoders. We replace the first and last layer in the encoder and decoder, respectively, with a decomposed layer (Decomposed model) or a separate linear layer for each subject (Subject model). Each autoencoder is trained and evaluated on three regions of interest (ROIs) and whole-brain data for the Sherlock and Forrest Gump datasets. The subject layer can not be evaluated on whole-brain data due to its exorbitant memory footprint, see Appendix~\ref{app:scaling}. For example, the Forrest Gump dataset contains 16 subjects, and each volume is 441100 voxels. With 256 as the MLP's hidden size, the total number of parameters for the subject layers in the encoder and decoder is around 3.6B. For the same setting, the Decomposed layer, as explained in Section~\ref{sec:method}, only uses 226M parameters, and the Group model uses 225M parameters. In all cases, as shown in Figure~\ref{fig:classification}, both our Subject and Decomposed models outperform the Group model, and in most cases this difference is significant. Moreover, in all cases, our models outperform the MRMD-AE. Using whole-brain data additionally boosts the classification accuracy for each context label, especially for the Sherlock dataset. This highlights the benefit of working with whole-brain data instead of focusing on a single ROI. For whole-brain data, the Decomposed model significantly improves performance for each dataset and type of context label. These results indicate that linear subject-specific spatial maps improve non-linear latent manifold learning.

\begin{figure}[H]
    \centering
    \includegraphics[width=\textwidth]{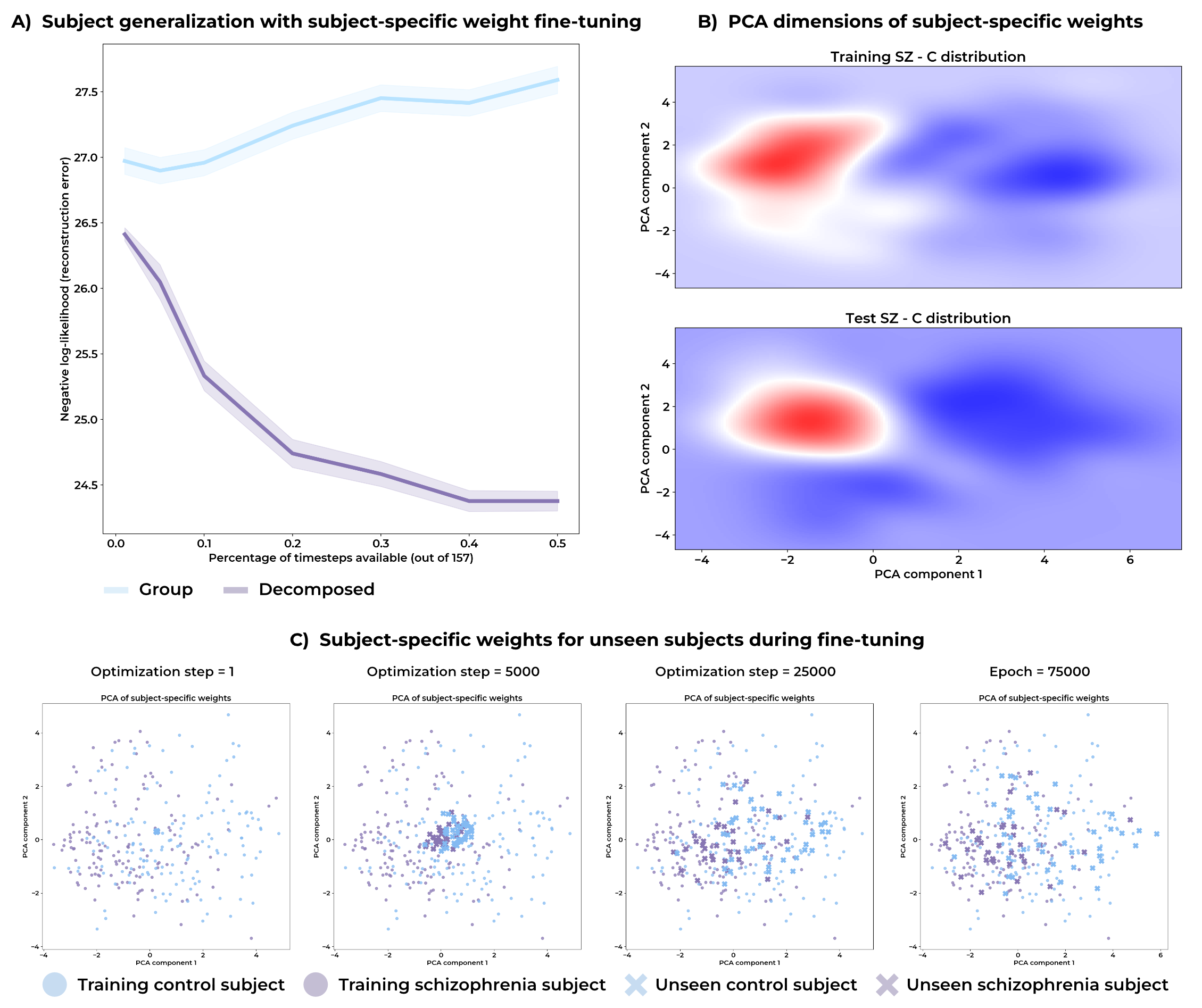}
    \caption{\textbf{Subfigure a}: We evaluate reconstruction performance as a function of the percentage of data we fine-tune the subject-specific weights of unseen subjects on. The total number of timesteps is 157, and even with 1\% of those timesteps, our model outperforms the group layer in terms of reconstruction. During fine-tuning all layers in the model are frozen except for the subject-specific weights for unseen subjects. \textbf{Subfigure b}: Group distribution difference plots of subject-specific weights in a 2D PCA space. Individuals with schizophrenia are more concentrated in the left side of the plot, whereas control subjects are more concentrated towards the right side in the plot. These concentrations are also replicated for fine-tuned weights from unseen subjects. \textbf{Subfigure c}: A visualization of the subject-specific weight updates for unseen subjects (on the fBIRN rs-fMRI schizophrenia dataset) during fine-tuning. Over time, many of the unseen subject weights move to areas that contain higher densities of training subjects with that same diagnosis.}
    \label{fig:generalization}
\end{figure}

\subsection{Subject generalization and reliability}
\label{sec:subject-generalization}
An important benefit of our Decomposed model is that it scales well to a large number of subjects, see Appendix~\ref{app:scaling}. It is however both important that we can train the Decomposed model on a large set of subjects, but also that it generalizes well to new subjects To test this, we trained a variational autoencoder (VAE) with our Decomposed model on unstructured resting-state fMRI (rs-fMRI) data from control subjects and subjects diagnosed with schizophrenia. The ability of the Decomposed model to generalize to new subjects is evaluated by using 294 subjects in the training and validation set and keeping 74 unseen subjects in the test set. Since only the subject-specific weights are different for each subject, we can train the VAE on the 294 subjects in the training set, freeze all the weights in the VAE, and only update new subject-specific weights for unseen subjects. This process is visualized in Figure~\ref{fig:generalization}c. We take a small subset of the new subject's data and update the subject-specific weights for a number of iterations until the updates have converged. This means that the full model is not re-trained, and we only try to find the subject-specific weights for new subjects, a much more efficient way of adding new data into the model that does not cause catastrophic forgetting~\cite{french1999catastrophic}. After updating the subject-specific weights on a subset of the unseen subjects' data, we evaluate the VAE's performance on the unseen subjects' held-out set of data in Figure~\ref{fig:generalization}a. A larger subset of data leads to better reconstruction performance on the held-out set of data, but even with 1\% of the unseen subjects' data (1 TR), our model reconstructs the held-out 99\% of the timesteps better than the Group model.

\begin{figure}[H]
    \centering
    \includegraphics[width=\textwidth]{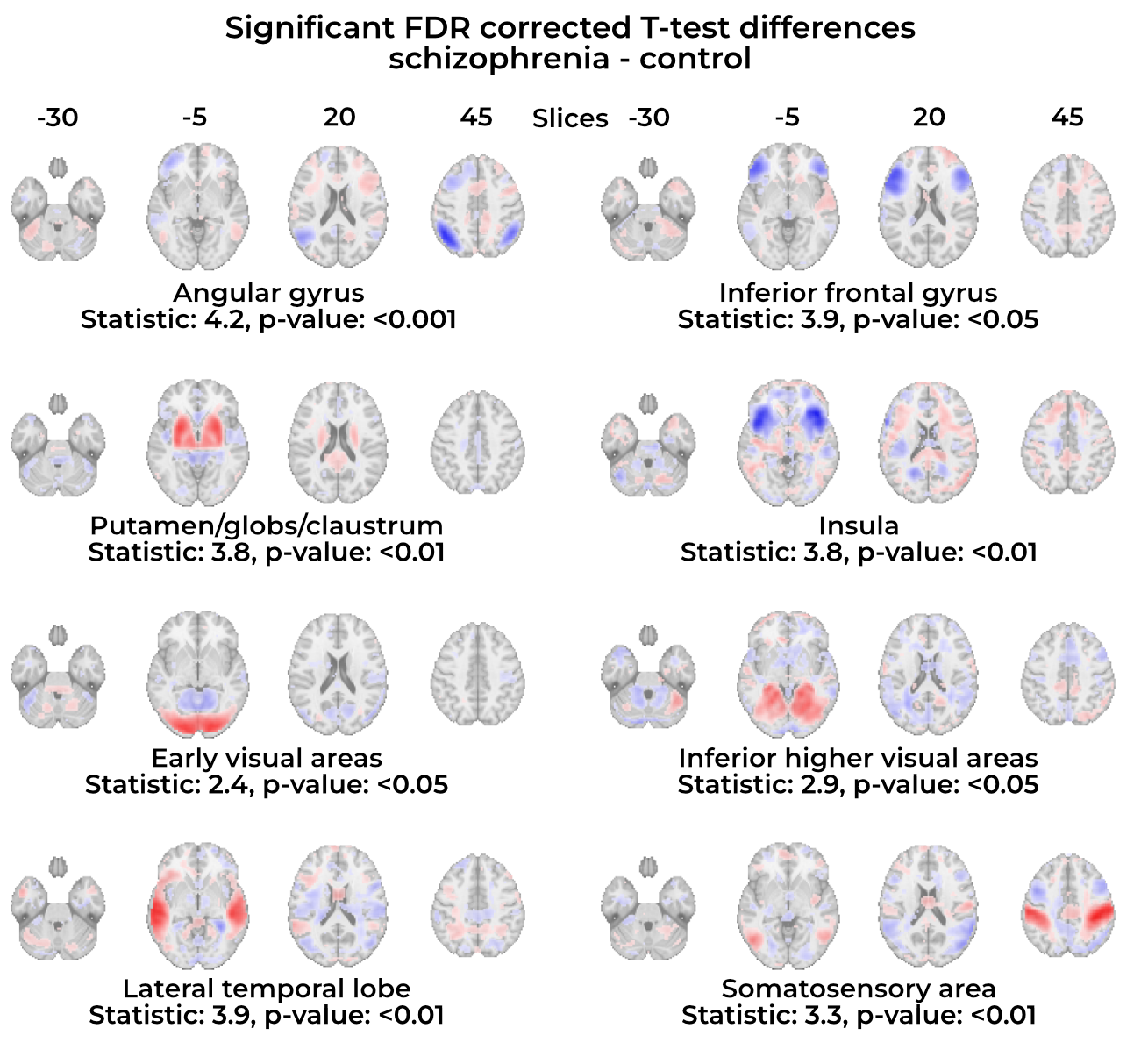}
    \caption{Significantly different brain activations between schizophrenia patients and control subjects as indicated by our Decomposed model. These brain regions are a selection from all the significant regions. We generally find significantly increased basal ganglia, visual, auditory, and somatosensory activations for schizophrenia patients. The inferior frontal gyrus, angular gyrus, and insula show significantly decreased activity however. The significance levels are corrected using the false discovery rate based on $64$ tests.}
    \label{fig:schizophrenia}
\end{figure}

Since the subject-specific weights generalize well in terms of reconstruction performance, we are also interested whether they capture clinically relevant subject differences. To evaluate whether the subject weights capture the class differences in the dataset, we perform a 20-fold classification task with the training subject weights and a radial basis function (RBF) support vector machine (SVM)~\cite{chang2011libsvm,scikit-learn}, which yields a classification accuracy of 79.16\%. This accuracy is rather high because our model does not have access to class labels during training. The differences between subject-specific weights are also visualized in Figure~\ref{fig:generalization}b, where schizophrenia subjects are concentrated towards the left side of the plot, whereas control subjects are concentrated towards the right side. This accuracy is impressive because our model did not have access to class labels during training, and was purely trained as an unsupervised model. In Figure~\ref{fig:generalization}b, the subject-specific weights of the unseen schizophrenia subjects are concentrated towards the left side of the plot, whereas the unseen control subjects are concentrated towards the right side. The group distribution differences between unseen and training subjects are largely similar, which indicates that the schizophrenia-specific features generalize to unseen schizophrenia subjects. Even with all other parameters in the neural network frozen, unseen subjects occupy similar areas of the subject-specific weight space as in the training set based on their diagnosis.

\subsection{Spatial subject-specific insights into schizophrenia}
\label{sec:schizophrenia-results}
To better understand what spatial differences the subject-specific weights capture for schizophrenia patients, we visualize a selection of areas with significant differences in Figure~\ref{fig:schizophrenia}. First, we generate subject-specific reconstructions for each latent dimension in our VAE model. To do this, we reconstruct the same series of latent points for each subject along the latent dimension. The only difference in the reconstruction for each subject is that we use a given subject's subject-specific weights. This ensures that any group difference is entirely due to the subject-specific weights. We run spatial Fast ICA~\cite{hyvarinen2000independent} on the subject-specific reconstructions to obtain $64$ sources. We then tested each source for significant group differences with a two-sided t-test between the schizophrenia and control entries of the whitened unmixing matrix. The p-values were corrected using the false discovery rate (FDR), and we selected $8$ significant sources for visualization in Figure~\ref{fig:schizophrenia}. 

The regions highlighted in Figure~\ref{fig:schizophrenia} are consistent with and extend previous work. Specifically, we find increased activity in the basal ganglia (putamen/globus/claustrum). The basal ganglia is known to be involved in motor function and emotion and motivation regulation. Lack of motivation and emotion regulation are symptoms related to psychosis and schizophrenia~\cite{medalia2010search,aleman2005strange}.  The differences in basal ganglia and the somatosensory area could also be consistent with the motor symptoms associated with schizophrenia~\cite{walther2012motor}. Another explanation for the increased activation in the somatosensory region is de-normalization due to psychotic treatment~\cite{c2013antipsychotic}. Reduced activation in the inferior frontal gyrus (IFG) has been linked to schizophrenia during a word serial position task~\cite{stevens1998cortical} and semantic coding~\cite{jeong2009functional}. The latter also found reduced functional connectivity in the language network that includes the IFG, and a connection between functional and anatomical connectivity abnormalities.  Generally, structural aberrations in the IFG are linked to schizophrenia, both in terms of tractographic abnormalities~\cite{kubicki2011stochastic} and reduced volume in the left IFG~\cite{schoretsanitis2019inferior}, which is also related to aggressive behavior, an externalizing symptom of schizophrenia. Alterations in the insula on the other hand are thought to be involved in internalizing symptoms such as hallucinations and the processing of both visual and auditory emotional information~\cite{wylie2010role}. In Figure~\ref{fig:schizophrenia} we see decreased insula activity and increased activity in early visual, higher visual, and auditory areas for schizophrenia patients. The increased activity in the visual and auditory areas could be counteracting the reduced insula activity or indicative of visual~\cite{butler2008visual} and auditory processing~\cite{javitt2015auditory} deficits, although this requires more research. Lastly, previous work~\cite{niznikiewicz2000abnormal} found the left angular gyrus to be smaller than the right angular gyrus, a reversed asymmetry with respect to control subjects, in schizophrenia subjects. In Figure~\ref{fig:schizophrenia} we see a larger reduction  in angular gyrus activation on the left side than on the right side, which could be due to the reported asymmetry in angular gyrus volume. These findings indicate the unique ability of our framework to capture subject-specific spatial variation in a way that is interpretable and psychiatrically relevant. Moreover, the fact that our model can learn relevant individualized spatial maps for around $300$ subjects completely end-to-end demonstrates its potential applicability to a wide range of neuroimaging datasets.

\section{Discussion}\label{sec:discussion}
Recent work has shown the importance of subject-specific non-linear latent manifolds for naturalistic fMRI data~\cite{busch2023multi}. This approach was subsequently extended to incorporate environmental factors and predict cognitive measures~\cite{busch2024manifold}. However, an important downside to individualized latent manifolds is that they can complicate group comparisons on unstructured data, i.e. data that is not temporally aligned among subjects. This complicates learning latent manifolds from rs-fMRI data, a common type of unstructured fMRI data that is often used to study mental disorders. Although there are approaches that have focused on learning non-linear latent group manifolds for fMRI data~\cite{gao2021nonlinear}, there is a lack of approaches that combine both subject-specific and group approaches for learning non-linear latent manifolds.

In Section~\ref{sec:reconstruction-results} we first show how mixing subject-specific and group weights allows us to learn better latent manifolds. Both in terms of reconstructing unseen timepoints, and in terms of classifying events in the movie subjects are watching. By keeping the subject-specific weights linear with respect to the input for both our Subject and Decomposed models, subject differences are inherently interpretable. Although subject-specific linear maps are used in shared response modeling~\cite{chen2015reduced}, and fMRI decoding~\cite{scotti2024mindeye2}, our approach is the first that generalizes linear subject-specific approaches to unstructured data and manifold learning, in a way that generalizes to a large number of subjects. Specifically, we introduce a Decomposed version of the subject-specific maps that scales well in the number of parameters with increasing input and dataset sizes, see Appendix~\ref{app:scaling}. Since the Decomposed model scales well with input size, we can apply it to voxelwise whole-brain data. The Decomposed model significantly improves performance compared to the commonly used Group model on whole-brain data, and also robustly improves performance on the Sherlock data with respect to methods trained only on ROIs. This both confirms the importance of using whole-brain data, and the ability of our Decomposed model to scale to large input sizes and effectively utilize the additional information in whole-brain data.

To evaluate how well the Decomposed model scales, we apply our model on unstructured rs-fMRI data in Section~\ref{sec:subject-generalization}, and test how well it generalizes to new subjects. We show that our model can generalize to new subjects by only fine-tuning the subject-specific weights and keeping all other weights frozen, preventing catastrophic forgetting~\cite{french1999catastrophic}. We find that the Decomposed model easily generalizes to new subjects-- even if only a single TR is available for each subject. Moreover, we find that the subject-specific weights capture group differences between control subjects and subjects diagnosed with schizophrenia. These differences are easily seen when we project the subject-specific weights onto their first two PCA components in Figure~\ref{fig:generalization}b. When we infer subject-specific weights of unseen subjects, we show that the distribution of control subjects and subjects diagnosed with schizophrenia is similar when projected onto the same PCA components. This both indicates that our model can infer psychiatrically relevant subject-specific differences in an unsupervised way, and that these subject-specific differences generalize to new subjects.

To further explore these differences between control subjects and subjects diagnosed with schizophrenia, we visualize the spatial differences captured by the subject-specific weights in Figure~\ref{fig:schizophrenia}. We find significant differences in a variety of regions, including increases in activations in the basal ganglia, early and late visual areas, the lateral temporal lobe, and the somatosensory areas. We hypothesize that the increases in the basal ganglia and somatosensory areas could be linked to motor symptoms in schizophrenia, and that the increases in the lateral temporal lobe and visual areas could either be compensatory activations or related to visual and auditory processing issues in schizophrenia. We also find asymmetrical decreases in the angular gyrus, and decreases in the insula and inferior frontal gyrus (IFG). We speculate that the changes in the IFG could be linked to externalizing symptoms, whereas the decrease in activation in the insula could be linked to internalizing symptoms. The asymmetric decrease in the angular gyrus aligns with previous work into schizophrenia. Although these findings require further study, they do indicate the unique ability of our framework to capture subject-specific spatial variation in a way that is interpretable and psychiatrically relevant.

The models we propose here are general and can easily be used with different neural network architectures, with other neuroimaging data, and even with other hierarchically structured scientific data. Future work could focus on incorporating dynamics more explicitly in the model architecture. It is also possible to use the (Decomposed) subject-specific weights in other parts of the neural network architecture. This would allow the model to have subject-specific temporal differences, instead of spatial differences as in this work, for example. Moreover, with a growing interest in fMRI foundation models~\cite{caro2023brainlm,dong2024brain,park2025foundational}, and the incorporation of session or unit-specific embeddings in other neuroscientific foundation models~\cite{azabou2023unified}, the incorporation of subject-specific weights in fMRI foundation models may drive further progress in the field.

\section{Methods}\label{sec:method}
\subsection{Subject-specific spatial maps}
In this work, a neuroimaging dataset ($\{\mathbf{X}_i\}_{i=1, ..., M}$) consists of $M$ subjects, and each subject's data $\mathbf{X}_i \in \mathbb{R}^{T\times N}$ has $N$ voxels and $T$ timesteps. An immediate problem arises since  generally $N >> M$, i.e. $N$ is on the order of $10^5$, whereas many studies only have $100-1000$ subjects ($M$). Additionally, approaches like SRM~\cite{chen2015reduced} assume the following underlying generative model assumption: $\mathbf{X}_i = \mathbf{S} \mathbf{W}_i  + \mathbf{E}_i$, where $\mathbf{W}_i \in \mathbb{R}^{d\times N}$ is a subject-specific spatial spatial activation map and $d$ the shared response's assumed low-dimensional latent space, $\mathbf{S} \in \mathbb{R}^{T \times d}$ is the shared response that is assumed to be the same across subjects, and $\mathbf{E}_i \in \mathbb{R}^{T \times N}$ is a subject-specific noise term. Given the success of approaches like SRM in assuming a separate spatial activation map for each subject, we develop a framework that generalizes learning subject-specific spatial maps with unstructured generative models. The main issue that arises is that the number of spatial map parameters scales with $N\times d\times M$, whereas the dataset size scales with $T\times M$. Although $d << N$, the extremely large size of $N$ makes it hard to find subject-specific maps for smaller datasets, i.e. where $T$ is relatively small. Specifically, there are too many parameters to fit and too few data examples for fitting to the parameters. In this work, we show that a more general generative model is helpful, and also propose an efficient way of estimating subject-specific spatial maps for datasets with few timesteps ($T$).

First we introduce the new generative model. Generally, neuroimaging data is assumed to lie on a low-dimensional manifold, so although the number of voxels $N$ in a dataset is large, much of those voxels can be explained by a smaller number of factors, e.g. regions of interests (ROIs) in an atlas. In our case, we want to learn both the mapping to and from that low-dimensional manifold with a neural network. Given the ability of neural networks to approximate any function according to the universal approxiation theorem, they can theoretically approximate any potential low-dimensional manifold of neuroimaging data. We first generalize the use of subject-specific spatial maps to autoencoders. Starting with the simplest autoencoder, PCA is an optimal solution to a linear autoencoding problem. Recall that $\mathbf{X}_i \in \mathbb{R}^{T\times N}$ is our neuroimaging signal for each subject $i$. We want to map each subject's timeseries onto a low-dimensional manifold so we can analyze its temporal evolution on this manifold. For example, in computational neuroscience, we may constrain the temporal evolution to be modeled by a discrete dynamical system~\cite{lfads}. If we assume that the dimensions of our neural manifold need to be orthogonal and linearly related to the original data, we can solve the linear autoencoding problem using subject-specific PCA, and obtain an optimal, orthogonal $\mathbf{W}_i$. Each $\mathbf{W}_i$ will be different for each subject since we solve PCA independently for each subject, complicating group comparisons. Moreover, solving this optimization problem with PCA leads to two more issues. First, the optimization problem is inflexible and does not allow for non-linear manifolds or manifolds with specific constraints. Second, each neural manifold is now different for each subject, which complicates group analysis. We would preferably: take subject-specific differences into account when mapping $\mathbf{X}_i$ into some neural feature space, be able to compare subjects on the same/similar neural manifold, and allow for an extremely flexible mapping onto this manifold so we can create complex constraints for the manifold.

To allow for extremely flexible mappings, we will first discuss how the generative model is described with a neural network. Let us define a similar optimization problem, but use a subject-specific neural-network-based autoencoder.
\begin{equation}
    \argmin_{\theta_i, \phi_i} \sum_{t=1}^{T} \lVert \mathbf{X}_{(i, t)} - g_{\phi_i}(f_{\theta_i}(\mathbf{X}_{(i, t)})) \rVert_{2}^{2}
\end{equation}
where $f_{i, \theta_i}: \mathbb{R}^{N} \to \mathbb{R}^d$ is an encoder,  $g_{i, \phi_i}: \mathbb{R}^{d} \to \mathbb{R}^N$ is a decoder, and $\theta_i$ and $\phi_i$ are the subject-specific parameters we are optimizing over. Defining a separate encoder and decoder for each subject ($i$) can easily lead to overfitting and does not result in a shared latent space across subjects. Instead, we can share the encoder and decoder across subjects, but find a unique first (and last) linear layer for each subject as follows.
\begin{subequations}
\label{eq:subject-autoencoder}
\begin{align}
    f_{i, \theta}(\mathbf{X}_i) &= f_{\theta}(\mathbf{X}_i \mathbf{W}^{\text{enc}}_i) = z_{(i, t)} \\
    g_{i, \phi}(z_{(i, t)}) &= \mathbf{W}^{\text{dec}}_i g_{\phi}(z_{(i, t)}) = \hat{\mathbf{X}}_{(i, t)}
\end{align}
\end{subequations}
where $\mathbf{W}^{\text{enc}}_{i} \in \mathbb{R}^{N \times L}$, $\mathbf{W}^{\text{dec}}_{i} \in \mathbb{R}^{L \times N}$, $f_{\theta}: \mathbb{R}^{L} \to \mathbb{R}^{d}$, $g_{\theta}: \mathbb{R}^{d} \to \mathbb{R}^{L}$, and L is a hidden size in the network, where $L << N$. All the parameters $\theta$ and $\phi$ are now shared across the subjects, so they map each subject's temporal trajectory onto the same group neural manifold, but we still take individual differences into account using the first linear layer in the encoder, and the last linear layer in the decoder. Note that although we defined a reconstruction optimization problem, we can essentially define any other objective similarly, such as a classification or self-supervised objective, and optimize over the parameters and each subject-specific spatial map using gradient descent.

The approach in Equation\ref{eq:subject-autoencoder} would be quite attractive if we had enough data to train these models. Finding the parameters for the subject-specific matrices leads to a huge parameter space. For 1200 subjects, with 150k voxels per volume, and $L = 10$ we would obtain a total of $2 \times 1200 \times 10 \times 150000 = 3.6$B parameters, just for the subject-specific spatial maps. Moreover, from an engineering perspective, it is prohibitively expensive to store this many parameters in memory while training a neural network.

\subsection{Efficient estimation of spatial maps}
To overcome the data efficiency issue, we assume that each of these subject-specific spatial maps share structure with each other. We know this assumption is reasonable based on the similarity of individual spatial maps of resting networks from ICA~\cite{calhoun2012multisubject}. Additionally, although important sub-areas for cognitive functions, e.g. motor function for left hand tapping, may have slightly different subject-to-subject shapes, much of the structure is shared across subjects. To efficiently estimate subject-specific spatial maps, we propose a new method that leverages this shared structure.

By assuming some shared structure in the subject-specific maps, we do not have to learn a completely independent matrix for each subject. Instead, we can decompose each subject-specific spatial map $W_i \in \mathbb{R}^{L\times N}$ as follows.
\begin{equation}
    \mathbf{W}_i = \mathbf{U}_i\mathbf{S}_i\mathbf{V}^{T}_i
\end{equation}
Where $\mathbf{U}_i \in \mathbb{R}^{L \times R}$, $\mathbf{S}_i \in \mathbb{R}^{R \times R}$, $\mathbf{V}^{T}_i \in \mathbb{R}^{R \times N}$, $\mathbf{S}_i$ is a diagonal matrix with singular values $\mathbf{s}_i$ on its diagonal, $\mathbf{U}_i$ and $\mathbf{V}^{T}_i$ have orthonormal columns, and $R$ is the rank and in our case equal to $min(N, L)$. Instead of optimizing each $\mathbf{U}_i$ and $\mathbf{V}_i$ separately, we can assume that these matrices are shared between subjects and that only singular values vary across subjects as follows.
\begin{equation}
    \mathbf{W}_i = \mathbf{U}\mathbf{S}_i\mathbf{V}^{T}
\end{equation}
Then, instead of optimizing over a total of $NdM$ parameters for the subject-specific matrices $\{\mathbf{W}_i\}_{i=1}^{N}$, we can reduce the number of parameters to $LM + NL + L^2$. Using the numbers in the previous example, we go from 3.6B parameters to $2 * (10 * 150,000 + 1200 * 10 + 100) = 3.02$M, a reduction of 3 orders of magnitude. Moreover, using a group spatial map would result in $2 * (10 * 150,000) = 3$M parameters, which is only slightly fewer parameters than our new reparameterized subject-specific spatial maps. A more in-depth look at parameter scaling is shown in Appendix~\ref{app:scaling}. Our new decomposition is thus much more efficient because $L << N$ and parameterizing the singular values across subjects means the number of subject-specific parameters scale with $L$ instead of $N$. With this new decomposed subject-specific spatial map, we can replace $\mathbf{X}_{(i, t)} \mathbf{W}^{T}_{i}$ with two linear layers and an element-wise multiplication of the subject-specific singular values. In our implementation we only enforce orthogonality on the $L \times L$ matrix, which is different depending on whether we are talking about the encoder or decoder, because it is too computationally expensive to enforce orthogonality on the $N \times L$ matrix.

\subsection{Data}
\subsubsection{Simulated data}
In this study, we use the Half Moons dataset, as available through scikit-learn~\cite{scikit-learn} for the simulation results as discussed in Section~\ref{sec:simulation-results}. The data parameters include a noise level of $0.1$, and we simulate $1000$ samples with $42$ as the random state. These samples are used as a simulation of fMRI timesteps. Then, to obtain different subjects, we use the $1000$ samples, and rotate them using a random angle drawn from a uniform distribution $\theta \thicksim \mathit{U} \left[-2\pi, 2\pi \right]$. Since the Half Moons dataset is two-dimensional, the rotation is then performed by multiplying each sample with the following rotation matrix.
\begin{equation}
   X_{i,t} = \begin{bmatrix}
            \cos(\theta_i) & \sin(\theta_i) \\
            -\sin(\theta_i) & \cos(\theta_i)
   \end{bmatrix} Z_{t}
\end{equation}
Where $t$ indexes the sample from the Half Moons dataset, $i$ indexes the subjects, and $z$ are the Half Moons dataset samples. The rotations result in $100$ subjects with $1000$ samples for each subject. Since all subjects are included in the training, validation, and test set, the Half Moons samples are split into a training, validation, and test set. The test set is $80$\% of the Half Moon samples, and the training and validation set are $90$\% and $10$\% splits of the remaining samples, respectively. 

\subsubsection{Naturalistic stimulus fMRI datasets}
To evaluate how well the latent representations learned by the autoencoder version of our model align with cognitive processes, we used two structured naturalistic stimulus datasets. Although the model is not trained in a way that takes the structured information of the dataset into account, the datasets have labels for each timestep, which we use for a classification task in Section~\ref{sec:reconstruction-results}. We also use the datasets to evaluate how well the autoencoder version of our model can reconstruct unseen fMRI timesteps in Section~\ref{sec:reconstruction-results}.

\paragraph{Sherlock} The dataset is described in full in the accompanying publication~\cite{chen2017shared}. The data was accessed in this work through DataSpace at the following link
\url{http://arks.princeton.edu/ark:/88435/dsp01nz8062179}. We used data from 16 of the 17 subjects, since there is missing data for subject 5. The scans for each subject have 1976 repetition times (TRs).

\paragraph{StudyForrest} The dataset is dscribed in full in the accompanying publication~\cite{hanke2014high}. The data was accessed in this work through Datalad at the following link \url{https://github.com/psychoinformatics-de/studyforrest-data}. We used 7T linearly anatomically aligned data from 16 out of 20 subjects, the data from subjects 04, 10, 11, and 13 were not used due to data issues described in the original dataset paper~\cite{hanke2014high}. The scans for each subject have 3541 repetition times (TRs).

\paragraph{Classification labels} To make comparisons to the MRMD-AE as similar as possible, we use labels made public by one of the co-authors for a more recent publication~\cite{busch2023multi} at the following link \url{https://github.com/ericabusch/tphate_analysis_capsule}. 

\subsubsection{Unstructured resting-state fMRI data of schizophrenia patients}
The unstructured rs-fMRI dataset that is used to evaluate subject generalization in Section~\ref{sec:subject-generalization} and to evaluate subject differences between subjects with and without a schizophrenia diagnosis in Section~\ref{sec:schizophrenia-results}, is the function bioinformatic research network (fBIRN) phase III dataset~\cite{keator2016function}. The rs-fMRI scans in the dataset are collected at seven consortium sites (University of Minnesota, University of Iowa, University of New Mexico, University of North Carolina, University of California Los Angeles, University of California Irvine, and University of California San Francisco). Each site records each subject's diagnosis, age at the time of the scan, gender, illness duration, symptom scores, and current medication, when available. Subjects were only included if they were between $18$ and $65$ years of age, and their schizophrenia diagnosis was confirmed by trained raters using the Structured Clinical Interview for DSM-IV (SCID)~\cite{first2002structured}. All schizophrenia patients were on a stable dose of antipsychotic medication either typical, atypical, or a combination for at least two months. Moreover, each schizophrenia patient was clinically stable at the time of the scan. Control subjects were excluded based on current or past psychiatric illness. The psychiatric illness exclusion was based on the SCID assessment or in case a first-degree relative had an Axis-I psychotic disorder. Written informed consent was obtained under protocols approved by the Institutional Review Boards at each consortium site for all the subjects included in the full dataset.

\clearpage
\bibliography{library}

\begin{thebibliography}{10}

\bibitem{aleman2005strange}
A.~Aleman and R.~S. Kahn.
\newblock Strange feelings: do amygdala abnormalities dysregulate the emotional brain in schizophrenia?
\newblock {\em Progress in neurobiology}, 77(5):283--298, 2005.

\bibitem{allen2012capturing}
E.~A. Allen, E.~B. Erhardt, Y.~Wei, T.~Eichele, and V.~D. Calhoun.
\newblock Capturing inter-subject variability with group independent component analysis of fmri data: a simulation study.
\newblock {\em Neuroimage}, 59(4):4141--4159, 2012.

\bibitem{azabou2023unified}
M.~Azabou, V.~Arora, V.~Ganesh, X.~Mao, S.~Nachimuthu, M.~Mendelson, B.~Richards, M.~Perich, G.~Lajoie, and E.~Dyer.
\newblock A unified, scalable framework for neural population decoding.
\newblock {\em Advances in Neural Information Processing Systems}, 36:44937--44956, 2023.

\bibitem{busch2024manifold}
E.~L. Busch, M.~I. Conley, and A.~Baskin-Sommers.
\newblock Manifold learning uncovers nonlinear interactions between the adolescent brain and environment that predict emotional and behavioral problems.
\newblock {\em Biological Psychiatry: Cognitive Neuroscience and Neuroimaging}, 2024.

\bibitem{busch2023multi}
E.~L. Busch, J.~Huang, A.~Benz, T.~Wallenstein, G.~Lajoie, G.~Wolf, S.~Krishnaswamy, and N.~B. Turk-Browne.
\newblock Multi-view manifold learning of human brain-state trajectories.
\newblock {\em Nature computational science}, 3(3):240--253, 2023.

\bibitem{butler2008visual}
P.~D. Butler, S.~M. Silverstein, and S.~C. Dakin.
\newblock Visual perception and its impairment in schizophrenia.
\newblock {\em Biological psychiatry}, 64(1):40--47, 2008.

\bibitem{c2013antipsychotic}
C.~C~Abbott, A.~Jaramillo, C.~E~Wilcox, and D.~A~Hamilton.
\newblock Antipsychotic drug effects in schizophrenia: a review of longitudinal fmri investigations and neural interpretations.
\newblock {\em Current medicinal chemistry}, 20(3):428--437, 2013.

\bibitem{calhoun2012multisubject}
V.~D. Calhoun and T.~Adali.
\newblock Multisubject independent component analysis of fmri: a decade of intrinsic networks, default mode, and neurodiagnostic discovery.
\newblock {\em IEEE reviews in biomedical engineering}, 5:60--73, 2012.

\bibitem{calhoun2001method}
V.~D. Calhoun, T.~Adali, G.~D. Pearlson, and J.~J. Pekar.
\newblock A method for making group inferences from functional mri data using independent component analysis.
\newblock {\em Human brain mapping}, 14(3):140--151, 2001.

\bibitem{caro2023brainlm}
J.~O. Caro, A.~H. d.~O. Fonseca, C.~Averill, S.~A. Rizvi, M.~Rosati, J.~L. Cross, P.~Mittal, E.~Zappala, D.~Levine, R.~M. Dhodapkar, et~al.
\newblock Brainlm: A foundation model for brain activity recordings.
\newblock {\em bioRxiv}, pages 2023--09, 2023.

\bibitem{chang2011libsvm}
C.-C. Chang and C.-J. Lin.
\newblock Libsvm: a library for support vector machines.
\newblock {\em ACM transactions on intelligent systems and technology (TIST)}, 2(3):1--27, 2011.

\bibitem{charest2014unique}
I.~Charest, R.~A. Kievit, T.~W. Schmitz, D.~Deca, and N.~Kriegeskorte.
\newblock Unique semantic space in the brain of each beholder predicts perceived similarity.
\newblock {\em Proceedings of the National Academy of Sciences}, 111(40):14565--14570, 2014.

\bibitem{charest2015brain}
I.~Charest and N.~Kriegeskorte.
\newblock The brain of the beholder: honouring individual representational idiosyncrasies.
\newblock {\em Language, Cognition and Neuroscience}, 30(4):367--379, 2015.

\bibitem{chen2017shared}
J.~Chen, Y.~C. Leong, C.~J. Honey, C.~H. Yong, K.~A. Norman, and U.~Hasson.
\newblock Shared memories reveal shared structure in neural activity across individuals.
\newblock {\em Nature neuroscience}, 20(1):115--125, 2017.

\bibitem{chen2015reduced}
P.-H.~C. Chen, J.~Chen, Y.~Yeshurun, U.~Hasson, J.~Haxby, and P.~J. Ramadge.
\newblock A reduced-dimension fmri shared response model.
\newblock {\em Advances in neural information processing systems}, 28, 2015.

\bibitem{dong2024brain}
Z.~Dong, R.~Li, Y.~Wu, T.~T. Nguyen, J.~Chong, F.~Ji, N.~Tong, C.~Chen, and J.~H. Zhou.
\newblock Brain-jepa: Brain dynamics foundation model with gradient positioning and spatiotemporal masking.
\newblock {\em Advances in Neural Information Processing Systems}, 37:86048--86073, 2024.

\bibitem{first2002structured}
M.~B. First, R.~L. Spitzer, M.~Gibbon, J.~B. Williams, et~al.
\newblock Structured clinical interview for dsm-iv-tr axis i disorders, research version, patient edition.
\newblock Technical report, SCID-I/P New York, NY, USA:, 2002.

\bibitem{french1999catastrophic}
R.~M. French.
\newblock Catastrophic forgetting in connectionist networks.
\newblock {\em Trends in cognitive sciences}, 3(4):128--135, 1999.

\bibitem{gao2021nonlinear}
S.~Gao, G.~Mishne, and D.~Scheinost.
\newblock Nonlinear manifold learning in functional magnetic resonance imaging uncovers a low-dimensional space of brain dynamics.
\newblock {\em Human brain mapping}, 42(14):4510--4524, 2021.

\bibitem{han2019variational}
K.~Han, H.~Wen, J.~Shi, K.-H. Lu, Y.~Zhang, D.~Fu, and Z.~Liu.
\newblock Variational autoencoder: An unsupervised model for encoding and decoding fmri activity in visual cortex.
\newblock {\em NeuroImage}, 198:125--136, 2019.

\bibitem{hanke2014high}
M.~Hanke, F.~J. Baumgartner, P.~Ibe, F.~R. Kaule, S.~Pollmann, O.~Speck, W.~Zinke, and J.~Stadler.
\newblock A high-resolution 7-tesla fmri dataset from complex natural stimulation with an audio movie.
\newblock {\em Scientific data}, 1(1):1--18, 2014.

\bibitem{haxby2020hyperalignment}
J.~V. Haxby, J.~S. Guntupalli, S.~A. Nastase, and M.~Feilong.
\newblock Hyperalignment: Modeling shared information encoded in idiosyncratic cortical topographies.
\newblock {\em elife}, 9:e56601, 2020.

\bibitem{huang2017modeling}
H.~Huang, X.~Hu, Y.~Zhao, M.~Makkie, Q.~Dong, S.~Zhao, L.~Guo, and T.~Liu.
\newblock Modeling task fmri data via deep convolutional autoencoder.
\newblock {\em IEEE transactions on medical imaging}, 37(7):1551--1561, 2017.

\bibitem{huang2022learning}
J.~Huang, E.~Busch, T.~Wallenstein, M.~Gerasimiuk, A.~Benz, G.~Lajoie, G.~Wolf, N.~Turk-Browne, and S.~Krishnaswamy.
\newblock Learning shared neural manifolds from multi-subject fmri data.
\newblock {\em 2022 IEEE 32nd International Workshop on Machine Learning for Signal Processing (MLSP)}, pages 01--06, 2022.

\bibitem{hyvarinen2000independent}
A.~Hyv{\"a}rinen and E.~Oja.
\newblock Independent component analysis: algorithms and applications.
\newblock {\em Neural networks}, 13(4-5):411--430, 2000.

\bibitem{iyer2022focal}
K.~K. Iyer, K.~Hwang, L.~J. Hearne, E.~Muller, M.~D’Esposito, J.~M. Shine, and L.~Cocchi.
\newblock Focal neural perturbations reshape low-dimensional trajectories of brain activity supporting cognitive performance.
\newblock {\em Nature communications}, 13(1):4, 2022.

\bibitem{javitt2015auditory}
D.~C. Javitt and R.~A. Sweet.
\newblock Auditory dysfunction in schizophrenia: integrating clinical and basic features.
\newblock {\em Nature Reviews Neuroscience}, 16(9):535--550, 2015.

\bibitem{jeong2009functional}
B.~Jeong, C.~G. Wible, R.-I. Hashimoto, and M.~Kubicki.
\newblock Functional and anatomical connectivity abnormalities in left inferior frontal gyrus in schizophrenia.
\newblock {\em Human brain mapping}, 30(12):4138--4151, 2009.

\bibitem{keator2016function}
D.~B. Keator, T.~G. van Erp, J.~A. Turner, G.~H. Glover, B.~A. Mueller, T.~T. Liu, J.~T. Voyvodic, J.~Rasmussen, V.~D. Calhoun, H.~J. Lee, et~al.
\newblock The function biomedical informatics research network data repository.
\newblock {\em Neuroimage}, 124:1074--1079, 2016.

\bibitem{kim2021representation}
J.-H. Kim, Y.~Zhang, K.~Han, Z.~Wen, M.~Choi, and Z.~Liu.
\newblock Representation learning of resting state fmri with variational autoencoder.
\newblock {\em NeuroImage}, 241:118423, 2021.

\bibitem{kubicki2011stochastic}
M.~Kubicki, J.~L. Alvarado, C.-F. Westin, D.~F. Tate, D.~Markant, D.~P. Terry, T.~J. Whitford, J.~De~Siebenthal, S.~Bouix, R.~W. McCarley, et~al.
\newblock Stochastic tractography study of inferior frontal gyrus anatomical connectivity in schizophrenia.
\newblock {\em Neuroimage}, 55(4):1657--1664, 2011.

\bibitem{medalia2010search}
A.~Medalia and J.~Brekke.
\newblock In search of a theoretical structure for understanding motivation in schizophrenia.
\newblock {\em Schizophrenia bulletin}, 36(5):912--918, 2010.

\bibitem{nastase2019measuring}
S.~A. Nastase, V.~Gazzola, U.~Hasson, and C.~Keysers.
\newblock Measuring shared responses across subjects using intersubject correlation, 2019.

\bibitem{niznikiewicz2000abnormal}
M.~Niznikiewicz, R.~Donnino, R.~W. McCarley, P.~G. Nestor, D.~V. Iosifescu, B.~O’Donnell, J.~Levitt, and M.~E. Shenton.
\newblock Abnormal angular gyrus asymmetry in schizophrenia.
\newblock {\em American Journal of Psychiatry}, 157(3):428--437, 2000.

\bibitem{lfads}
C.~Pandarinath, D.~J. O’Shea, J.~Collins, R.~Jozefowicz, S.~D. Stavisky, J.~C. Kao, E.~M. Trautmann, M.~T. Kaufman, S.~I. Ryu, L.~R. Hochberg, et~al.
\newblock Inferring single-trial neural population dynamics using sequential auto-encoders.
\newblock {\em Nature methods}, 15(10):805--815, 2018.

\bibitem{park2025foundational}
J.~Park, B.~Park, C.-B. Bang, J.~Choi, H.~Chung, B.-H. Kim, and J.~Lee.
\newblock A foundational brain dynamics model via stochastic optimal control.
\newblock {\em arXiv preprint arXiv:2502.04892}, 2025.

\bibitem{scikit-learn}
F.~Pedregosa, G.~Varoquaux, A.~Gramfort, V.~Michel, B.~Thirion, O.~Grisel, M.~Blondel, P.~Prettenhofer, R.~Weiss, V.~Dubourg, J.~Vanderplas, A.~Passos, D.~Cournapeau, M.~Brucher, M.~Perrot, and E.~Duchesnay.
\newblock Scikit-learn: Machine learning in {P}ython.
\newblock {\em Journal of Machine Learning Research}, 12:2825--2830, 2011.

\bibitem{richard2020modeling}
H.~Richard, L.~Gresele, A.~Hyvarinen, B.~Thirion, A.~Gramfort, and P.~Ablin.
\newblock Modeling shared responses in neuroimaging studies through multiview ica.
\newblock {\em Advances in Neural Information Processing Systems}, 33:19149--19162, 2020.

\bibitem{richard2019fast}
H.~Richard, L.~Martin, A.~L. Pinho, J.~Pillow, and B.~Thirion.
\newblock Fast shared response model for fmri data.
\newblock {\em arXiv preprint arXiv:1909.12537}, 2019.

\bibitem{schoretsanitis2019inferior}
G.~Schoretsanitis, K.~Stegmayer, N.~Razavi, A.~Federspiel, T.~J. M{\"u}ller, H.~Horn, R.~Wiest, W.~Strik, and S.~Walther.
\newblock Inferior frontal gyrus gray matter volume is associated with aggressive behavior in schizophrenia spectrum disorders.
\newblock {\em Psychiatry Research: Neuroimaging}, 290:14--21, 2019.

\bibitem{scotti2024mindeye2}
P.~S. Scotti, M.~Tripathy, C.~K.~T. Villanueva, R.~Kneeland, T.~Chen, A.~Narang, C.~Santhirasegaran, J.~Xu, T.~Naselaris, K.~A. Norman, et~al.
\newblock Mindeye2: Shared-subject models enable fmri-to-image with 1 hour of data.
\newblock {\em arXiv preprint arXiv:2403.11207}, 2024.

\bibitem{song2023large}
H.~Song, W.~M. Shim, and M.~D. Rosenberg.
\newblock Large-scale neural dynamics in a shared low-dimensional state space reflect cognitive and attentional dynamics.
\newblock {\em Elife}, 12:e85487, 2023.

\bibitem{stevens1998cortical}
A.~A. Stevens, P.~S. Goldman-Rakic, J.~C. Gore, R.~K. Fulbright, and B.~E. Wexler.
\newblock Cortical dysfunction in schizophrenia during auditory word and tone working memory demonstrated by functional magnetic resonance imaging.
\newblock {\em Archives of general psychiatry}, 55(12):1097--1103, 1998.

\bibitem{walther2012motor}
S.~Walther and W.~Strik.
\newblock Motor symptoms and schizophrenia.
\newblock {\em Neuropsychobiology}, 66(2):77--92, 2012.

\bibitem{wylie2010role}
K.~P. Wylie and J.~R. Tregellas.
\newblock The role of the insula in schizophrenia.
\newblock {\em Schizophrenia research}, 123(2-3):93--104, 2010.

\end{thebibliography}
\bibliographystyle{abbrv}

\begin{appendices}

\newpage
\section{Parameter scaling}
\label{app:scaling}%
\begin{figure}[H]
    \centering
    \includegraphics[width=\textwidth]{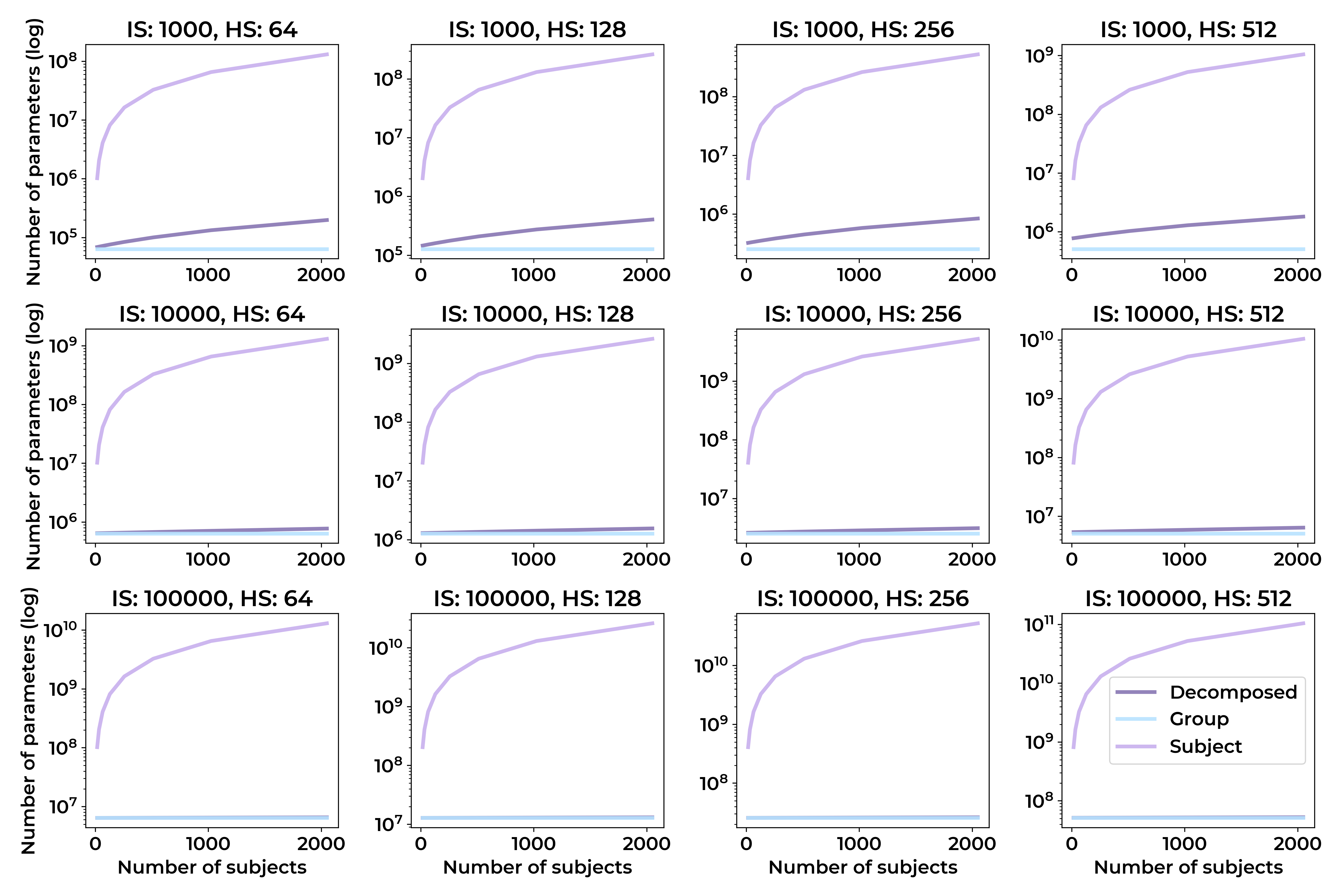}
    \caption{Each of these sub-figures indicates a different regime parameter scaling regime. \textbf{IS} is short for input size, and \textbf{HS} is short for hidden size. The number of subjects increase from left to right on the x-axis, and the number of parameters increase logarithmically on the y-axis. Each model is indicated by a different color, and the legend is shown in the bottom right plot. Note that in all cases, the Subject model leads to the largest number of parameters and does not scale well with the number of subjects. Moreover, especially in cases where the number of subjects is high and the input size is high the Decomposed model is relatively the closest to the Group model in terms of the number of the number parameters.}
    \label{fig:scaling}
\end{figure}

Neural networks with subject-specific spatial maps generally learn better non-linear manifolds as shown in Section~\ref{sec:reconstruction-results}. However, as the size of the input or the number of subjects in a dataset grow, the number of parameters of subject-specific maps also grow in a way that makes them cumbersome. How the number of parameters scale for each of the three main models we test in this work is shown in Figure~\ref{fig:scaling}. The Subject model can easily reach $10^{10}$ parameters for voxelwise data and a dataset with around $2000$ subjects. It is impractical to both train this many parameters with the limited size of neuroimaging datasets, and to keep the large number of parameters in memory during training. The Decomposed model, on the other hand, scales well with the input size and the number of subjects. The exact formulas that can be used to calculate the scaling laws are discussed below. 

Let IS be the input size, HS be the hidden size of the linear layer, NS be the number of subjects, and NP be the number of parameters. Then, for the Subject model we obtain the following equation.
\begin{equation}
    \text{NP} = \text{IS} \times \text{HS} \times \text{NS}
\end{equation}
The number of parameters for the Group model are calculated as follows.
\begin{equation}
    \text{NP} = \text{IS} \times \text{HS}
\end{equation}
Lastly, the number of parameters for the Decomposed model are calculated as follows.
\begin{equation}
    \text{NP} = \text{IS} \times \text{HS} + \text{HS} \times \text{HS} + \text{HS} \times \text{NS}
\end{equation}
Clearly, adding a new subject to the Decomposed model only increases the number of parameters by $\text{HS}$, which is much smaller than $\text{IS}$ in all relevant cases. This allows the number of parameters to scale well with a large number of subjects.

\end{appendices}
\clearpage
\end{document}